\documentclass{esannV2}

\usepackage{graphicx}
\usepackage[latin1]{inputenc}
\usepackage{amssymb,amsmath,array}
\usepackage{wrapfig}
\usepackage{hyperref}

\begin{document}

%***********************************************************************
% TITLE AND AUTHORS
%***********************************************************************

\title{The Dynamics of Quasiregular Neural Learning}

\author{
Matthia Sabatelli
\vspace{.3cm}\\
Department of Artificial Intelligence, University of Groningen\\
Groningen, The Netherlands\\
}

\maketitle

%***********************************************************************
% ABSTRACT
%***********************************************************************

\begin{abstract}
Many learning problems combine a dominant regularity with systematic
exceptions. Motivated by U-shaped learning in language acquisition, we study
this interaction in controlled quasiregular regression problems where regular
and exceptional solutions are explicitly known. Neural networks can partially
acquire exceptions, subsequently regress toward the dominant regularity, and
finally recover. This overregularization becomes substantially stronger when
exceptions are rare, despite their early acquisition, but does not emerge
equally across all regularities considered. Our results isolate a simple form
of competition between regularities and exceptions during neural learning.
\end{abstract}

%***********************************************************************
% INTRODUCTION
%***********************************************************************

\section{Introduction}
Learning is not always monotonic. A striking example comes from language
acquisition, where children learning the English past tense may initially
produce an irregular form correctly (e.g., \emph{go} $\rightarrow$
\emph{went}), later overregularize it (\emph{go} $\rightarrow$ \emph{goed}),
and eventually recover the correct form. This U-shaped trajectory played a
central role in debates over whether rule-like behaviour requires explicit
symbolic mechanisms or can emerge from distributed learning
\cite{rumelhart1986learning,marcus1992overregularization}. More recently,
similar dynamics have been revisited using modern neural architectures
\cite{kirov2018recurrent}.
One particularly interesting aspect of U-shaped learning is that exceptions
are not simply acquired late. Correct exceptional behaviour can appear before
overregularization and subsequently deteriorate as learning progresses
\cite{marcus1992overregularization}. The resulting trajectory therefore raises
a more general question than whether a learner can eventually represent both
rules and exceptions: how does learning a dominant regularity affect exception
information that has already been acquired?
This question is difficult to isolate in linguistic settings, where lexical
frequency, input representation, and properties of the language itself are
intertwined with the learning dynamics of interest. Here, we abstract away
from these factors and study \emph{quasiregular learning}: learning a
dominant regularity while accommodating systematic exceptions. While
quasiregular structure has previously been studied in distributed neural
models \cite{kim2013quasiregularity}, our focus is specifically on how the
interaction between regularities and exceptions unfolds during gradient-based
training.
We construct controlled regression problems in which the regular and
exceptional solutions are explicitly known. This allows us to go beyond
observing that exception performance temporarily deteriorates and ask whether
predictions specifically regress toward the competing regular solution. We
then use this setting to study how these dynamics depend on the relative
prevalence of exceptions and on the structure of the underlying regularity.

%***********************************************************************
% METHODS
%***********************************************************************

\section{Methods}
\label{sec:methods}

\subsection{Quasiregular learning problems}

\begin{wrapfigure}{r}{0.42\textwidth}
    \centering
    \vspace{-10pt}
    \includegraphics[width=\linewidth]{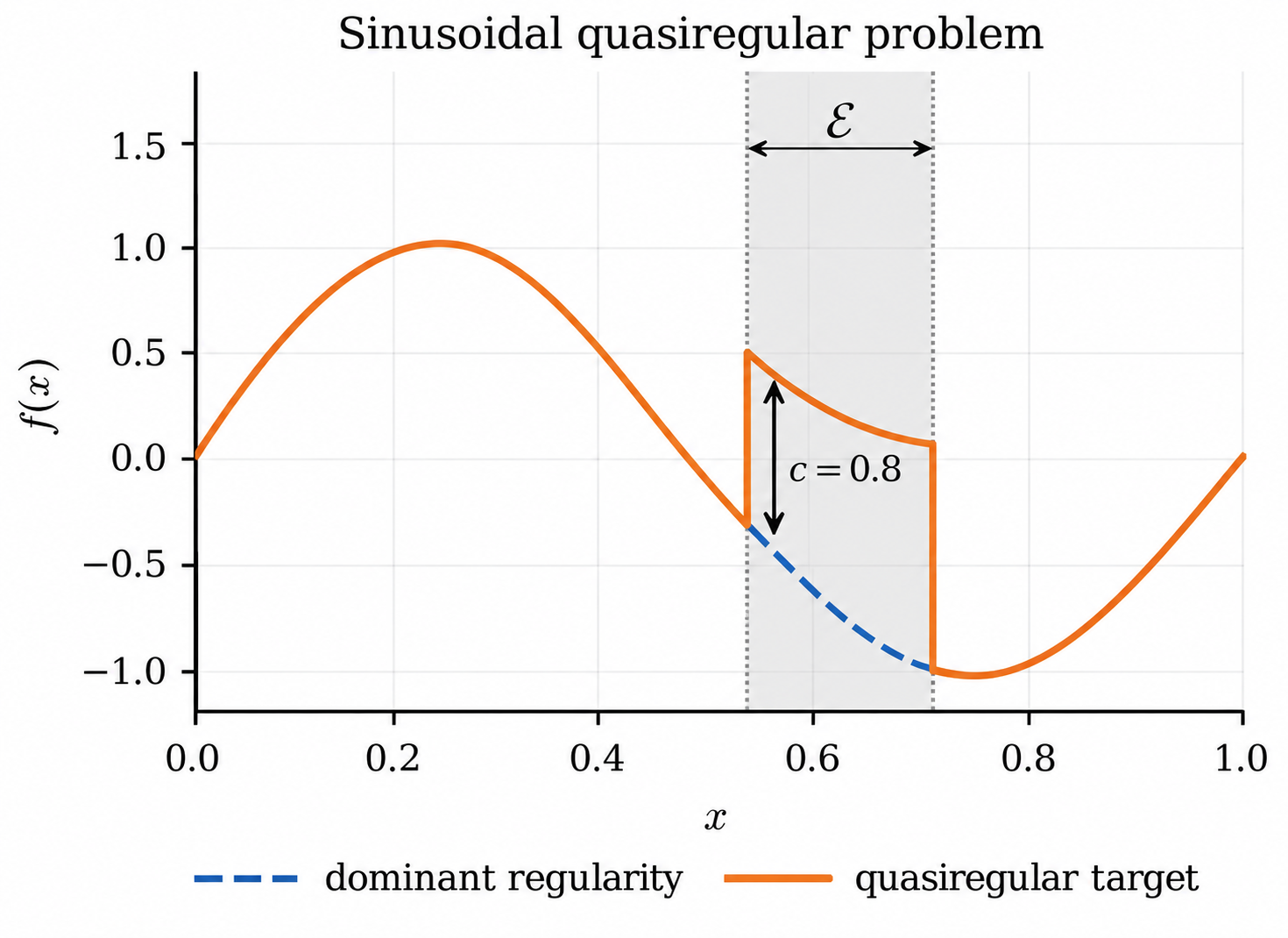}
    \caption{Sinusoidal quasiregular problem. The dominant regularity
    (dashed) is shifted by $c=0.8$ within the exception region
    $\mathcal{E}$.}
    \label{fig:quasiregular_problem}
    \vspace{-10pt}
\end{wrapfigure}

We study one-dimensional regression problems combining a dominant regularity
with systematic exceptions. Inputs $x\in[0,1]$ are associated with targets

\begin{equation}
    f(x)=
    \begin{cases}
        f_{\mathrm{reg}}(x)+c, & x\in\mathcal{E},\\
        f_{\mathrm{reg}}(x),   & \text{otherwise},
    \end{cases}
    \label{eq:quasiregular}
\end{equation}

where $f_{\mathrm{reg}}$ defines the dominant regularity,
$\mathcal{E}$ the exception region, and $c$ the exception offset. Our main
setting uses

\begin{equation}
    f_{\mathrm{reg}}(x)=\sin(2\pi x),
\end{equation}

with $\mathcal{E}=(0.55,0.70)$ and $c=0.8$
(Fig.~\ref{fig:quasiregular_problem}).

To investigate whether the resulting dynamics are specific to this
regularity, we additionally consider

\begin{align}
f_1(x) &= \sin(2\pi x)+0.3\sin(6\pi x),\\
f_2(x) &= \sin(2\pi x)+0.6\exp[-(x-0.3)^2/0.02],\\
f_3(x) &= 4(x-0.5)^3+0.5\sin(4\pi x).
\end{align}

The same exception region and offset are used for all four regularities,
allowing us to vary the dominant structure while leaving the exceptional
structure unchanged.

\subsection{Experimental setup and metrics}
We train a fully connected neural network with two hidden layers of 128 tanh
units using Adam ($\eta=10^{-3}$) and mean squared error for 1200 epochs.
Training sets contain 256 samples and evaluation is performed on 2000 uniformly
spaced test points.

In the main setting, inputs are sampled uniformly from $[0,1]$, resulting in
approximately $15\%$ exceptional training examples. Regular and exceptional
examples are therefore present together throughout training, and the training
distribution does not change over time. We repeat all experiments over 20
random seeds.
A useful property of this construction is that the regular and exceptional
solutions provide two known reference points. Within $\mathcal{E}$, the
regular solution predicts $f_{\mathrm{reg}}(x)$, while the correct exceptional
solution predicts $f_{\mathrm{reg}}(x)+c$. We can therefore measure where the
network lies relative to these solutions by subtracting the regular prediction
from its output. Let $\hat f_t(x)$ denote the network prediction for input
$x$ at epoch $t$. We define the mean exception offset as

\begin{equation}
    \Delta_t =
    \mathbb{E}_{x\in\mathcal{E}_{\mathrm{int}}}
    [\hat f_t(x)-f_{\mathrm{reg}}(x)],
    \label{eq:offset}
\end{equation}

where $\mathcal{E}_{\mathrm{int}}=(0.58,0.67)$ excludes points close to the
boundaries of the exception region, where the target transitions between
regular and exceptional solutions. Thus, $\Delta_t=0$ corresponds to the regular solution, whereas
$\Delta_t=c=0.8$ corresponds to the correct exceptional solution. A decrease
in $\Delta_t$ following an initial improvement in exception performance therefore
indicates that predictions are moving back toward the regular solution.
We additionally measure mean squared error
separately on regular and exceptional inputs.
We refer to a temporary regression of the exception offset toward the regular
solution following an initial improvement in exception performance as \emph{overregularization} \footnote{Here, overregularization refers to the
application of a dominant regularity to exceptional examples, following the
terminology of the language-learning literature; it is unrelated to
regularization in the statistical-learning sense.}.
To quantify its strength, we define the overregularization depth

\begin{equation}
    D=\Delta_{\mathrm{peak}}-\Delta_{\mathrm{trough}},
    \label{eq:depth}
\end{equation}

where $\Delta_{\mathrm{peak}}$ is the offset reached before regression and
$\Delta_{\mathrm{trough}}$ its subsequent minimum. Larger $D$ therefore
indicates stronger overregularization. Finally, to study the role of exception frequency, we vary the probability of
sampling training points from $\mathcal{E}$ as

\begin{equation}
p_{\mathrm{exc}}\in\{2.5,5,10,15,20,30,40\}\%,
\end{equation}

while keeping the exception region, offset, and total number of training
examples fixed. This changes the empirical prevalence of exceptions without
changing their location or target geometry.

%***********************************************************************
% RESULTS
%***********************************************************************

\section{Results}
\label{sec:results}

\subsection{Overregularization during learning}
\begin{figure}[hb!]
    \centering
    \includegraphics[width=0.88\textwidth]
    {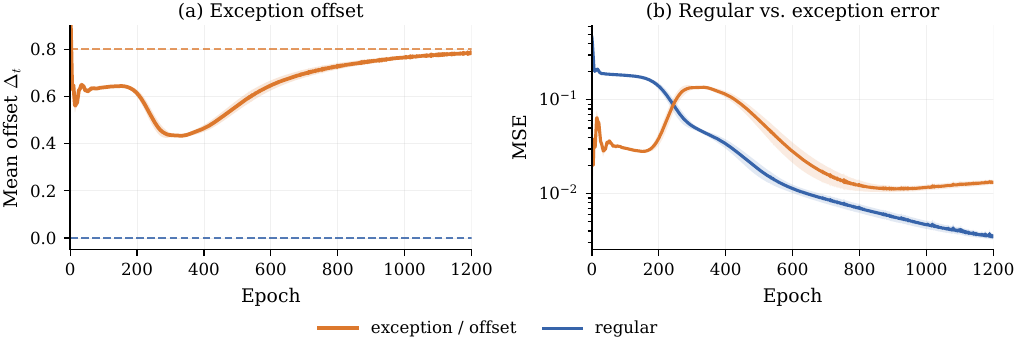}
    \caption{Learning dynamics on the sinusoidal quasiregular problem.
    Left: mean exception offset $\Delta_t$ over 20 seeds. Right: mean
    squared error on regular and exceptional inputs. Shaded regions denote
    95\% confidence intervals.}
    \label{fig:main_dynamics}
\end{figure}

We first ask how exception predictions evolve while the network learns the
dominant regularity. The error trajectories in the right panel of
Fig.~\ref{fig:main_dynamics} (note the logarithmic scale) show a clear
non-monotonic pattern for the exceptions. Early in training, exception error
decreases substantially, with MSE decreasing by approximately a factor of
1.5--2, demonstrating that the network initially improves its predictions on
exceptional examples. Exception performance then deteriorates sharply, with
exception error increasing several-fold while regular error continues to
decrease. Finally, exception error decreases again as training progresses.

The exception offset in the left panel reveals the direction of this
intermediate deterioration. As exception error increases, $\Delta_t$
decreases from approximately $0.65$ to $0.43$, indicating that exceptional
predictions shift toward the competing regular solution. The offset
subsequently recovers and approaches the correct exceptional value of $0.8$.
Together, the two panels reveal three stages of exception learning: initial
improvement, regression toward the regular solution, and recovery.

\subsection{Effect of exception prevalence}

Having established this trajectory, we next ask what determines its strength. A natural candidate is the relative frequency of the exceptions.

\begin{figure}[hb!]
    \centering
    \includegraphics[width=0.88\textwidth]
    {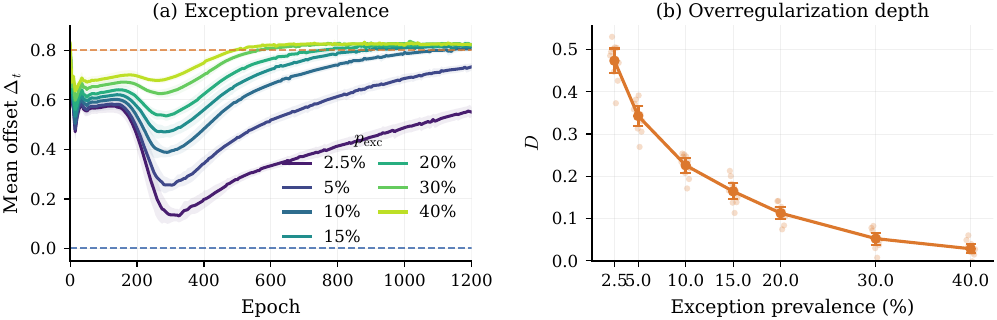}
    \caption{Effect of exception prevalence. Left: mean exception offset
    $\Delta_t$ for different exception frequencies. Right:
    overregularization depth $D$.}
    \label{fig:prevalence}
\end{figure}

Fig.~\ref{fig:prevalence} shows a clear dependence on exception prevalence.
When exceptions represent only $2.5\%$ of the training data, the network
initially moves substantially toward the exceptional solution, reaching an
offset above $0.5$. Its predictions then regress strongly toward the regular
solution before eventually recovering. As exception prevalence increases,
this intermediate regression becomes progressively smaller; at $30\%$ and
$40\%$, the U-shaped trajectory is strongly reduced.
This trend is summarized by the overregularization depth in the right panel
of Fig.~\ref{fig:prevalence}. The depth decreases from approximately $0.5$
when exceptions represent $2.5\%$ of the training data to almost zero at
$40\%$. Importantly, the rare-exception condition does not simply approach the
exceptional solution more slowly: even at $2.5\%$, predictions initially move
toward the exceptional reference before undergoing a pronounced regression.
Rarity therefore primarily affects the degree of subsequent regression: the
rarer the exceptions, the stronger their overregularization.

\subsection{Dependence on the regularity}

Exception prevalence is, however, not the only possible determinant of these dynamics.
We finally ask whether overregularization persists when the dominant
regularity itself is changed while keeping the exception structure fixed.

\begin{figure}[hb!]
    \centering
    \includegraphics[width=\linewidth]
    {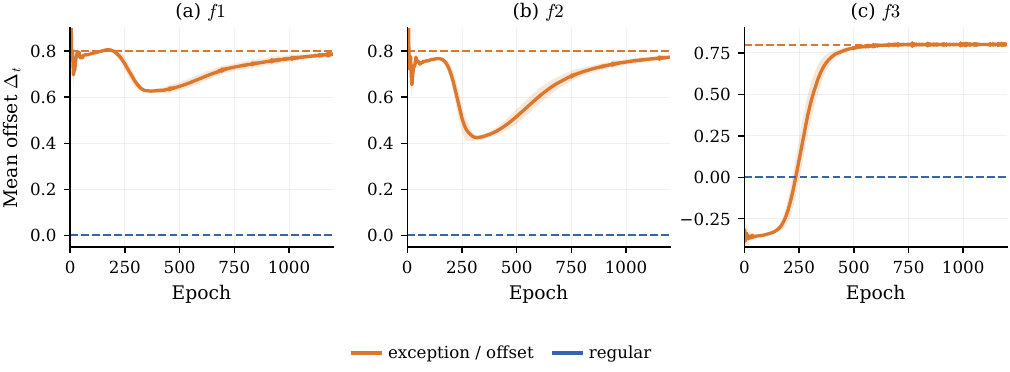}
    \caption{Exception learning dynamics for the three additional
    quasiregular problems.}
    \label{fig:additional_problem_results}
\end{figure}

The trajectories in Fig.~\ref{fig:additional_problem_results} show that the
behaviour observed in the sinusoidal problem is not unique to that particular
regularity. For both $f_1$ and $f_2$, the network initially moves toward the
exceptional solution, subsequently regresses toward the regular solution, and
finally recovers. The effect is particularly pronounced for $f_2$, where the
mean exception offset decreases from $0.768$ to $0.412$, corresponding to an
overregularization depth of $D=0.356$.
In contrast, $f_3$ shows a qualitatively different trajectory. Its exception
offset approaches the exceptional solution without a comparable intermediate
regression ($D=0.005$). Overregularization is therefore neither specific to
the sinusoidal problem nor an inevitable consequence of quasiregular
learning. %Whether it emerges, and how strongly, also depends on the structure
%of the dominant regularity.

%***********************************************************************
% DISCUSSION AND CONCLUSION
%***********************************************************************

\section{Discussion \& Conclusion}
\label{sec:conclusion}

We studied quasiregular learning in controlled regression problems where the regular and exceptional solutions are explicitly known. Across several problems, neural networks exhibit a characteristic U-shaped trajectory: they partially acquire exception information, regress toward the dominant regularity, and finally recover the exceptional solution. Crucially, this deterioration occurs while regular performance continues to improve. We have also shown that overregularization becomes substantially stronger as exceptions become rarer, despite exception information being acquired early in training. However, its absence for $f_3$ shows that rarity alone does not determine the trajectory: the structure of the dominant regularity also matters. Our setting relates to previous work on quasiregularity in distributed connectionist models \cite{kim2013quasiregularity}, while shifting the focus to the training-time interaction between regular and exceptional solutions in gradient-trained neural networks.
Our observations also connect to systematic biases in the order in which neural networks learn structure, such as the tendency to learn lower-frequency components faster \cite{rahaman2019spectral}. Our results show that such biases need not yield a monotonic ordering in which the dominant regularity is learned first and exceptions afterwards: exception information can emerge early, regress as the dominant regularity is learned, and subsequently recover. An important open question is what properties of a learning problem determine whether this competition produces overregularization. Characterizing these properties may connect quasiregular learning to broader accounts of neural-network learning dynamics, including the non-monotonic generalization dynamics of double descent.

\section*{Acknowledgements}

The authors would like to thank Gido van de Ven for insightful discussions
about this topic and Erwan Escudie for helpful feedback on preliminary versions
of the manuscript.

\paragraph{AI disclosure.}
ChatGPT (OpenAI) was used for spell-checking and to improve the clarity and
readability of the manuscript. All scientific ideas, experimental design,
analysis, and conclusions are those of the authors.

%***********************************************************************
% REFERENCES
%***********************************************************************

\bibliography{bibliography}
\bibliographystyle{plain}

\end{document}